\documentclass[letterpaper, 10 pt, journal, twoside]{IEEEtran}
\usepackage{amsmath,amsfonts}
\usepackage{algorithmic}
\usepackage{array}
\usepackage{textcomp}
\usepackage{stfloats}
\usepackage{url}
\usepackage{verbatim}
\usepackage{graphicx}
\def\BibTeX{{\rm B\kern-.05em{\sc i\kern-.025em b}\kern-.08em
		T\kern-.1667em\lower.7ex\hbox{E}\kern-.125emX}}
\usepackage{balance}

\usepackage{bm}
\usepackage[colorlinks=true,linkcolor=black,citecolor=green,urlcolor=blue]{hyperref}
\usepackage[capitalize]{cleveref}
\usepackage{algorithmic}
\usepackage[linesnumbered,boxed,ruled,commentsnumbered]{algorithm2e}
\usepackage{subfigure}
\usepackage{tikz}
\usepackage{underscore}
\usepackage{multirow}
\usepackage{booktabs}
\usepackage{amssymb}
\usepackage{makecell}
\usepackage{bbding}
\usepackage{cite}
\usepackage{float}
\usepackage{graphics} 
\usepackage{epsfig} 
\usepackage{txfonts}

\usepackage{soul}
\usepackage{color, xcolor}
\usepackage[framemethod=tikz]{mdframed}
\usepackage{xcolor}
\usepackage{framed}
\usepackage{lipsum}

\begin{document}
\title{BoW3D: Bag of Words for Real-Time Loop Closing \\ in 3D LiDAR SLAM}

\author{Yunge Cui\textsuperscript{1,2,3,4}, Xieyuanli Chen\textsuperscript{5}, Yinlong Zhang\textsuperscript{2,3,4}, Jiahua Dong\textsuperscript{2,3,4}, Qingxiao Wu\textsuperscript{1,2,3,4}, Feng Zhu\textsuperscript{1,2,3,4,$\dagger$}
  \thanks{Manuscript received: August 11, 2022; Revised: September 26, 2022; Accepted: October 29, 2022. 
  This paper was recommended for publication by Editor Javier Civera upon evaluation of the Associate Editor and Reviewers' comments.}%

	\thanks{$^{1}$Key Laboratory of Opto-Electronic Information Processing, Chinese Academy of Sciences, shenyang 110016, China} \thanks{$^{2}$Shenyang Institute of Automation, Chinese Academy of Sciences, Shenyang, China} \thanks{$^{3}$Institutes for Robotics and Intelligent Manufacturing, Chinese Academy of Sciences, Shenyang, China} \thanks{$^{4}$University of Chinese Academy of Sciences, Beijing, China} \thanks{$^{5}$The College of Intelligence Science and Technology, National University of Defense Technology, Changsha, China.} \thanks{$^\dagger$The corresponding author: Prof. Feng Zhu (Email: fzhu@sia.cn).}}

\maketitle

\markboth{IEEE Robotics and Automation Letters. Preprint Version. Accepted October, 2022}{Cui \MakeLowercase{\textit{et al.}}: BoW3D: Bag of Words for Real-Time Loop Closing in 3D LiDAR SLAM}

\begin{abstract}
Loop closing is a fundamental part of simultaneous localization and mapping (SLAM) for autonomous mobile systems. In the field of visual SLAM, bag of words (BoW) has achieved great success in loop closure. The BoW features for loop searching can also be used in the subsequent 6-DoF loop correction. However, for 3D LiDAR SLAM, the state-of-the-art methods may fail to effectively recognize the loop in real time, and usually cannot correct the full 6-DoF loop pose. To address this limitation, we present a novel \underline{B}ag \underline{o}f \underline{W}ords for real-time loop closing in \underline{3D} LiDAR SLAM, called BoW3D. Our method not only efficiently recognizes the revisited loop places, but also corrects the full 6-DoF loop pose in real time. BoW3D builds the bag of words based on the 3D LiDAR feature LinK3D, which is efficient, pose-invariant and can be used for accurate point-to-point matching. We furthermore embed our proposed method into 3D LiDAR odometry system to evaluate  loop closing performance. We test our method on public dataset, and compare it against other state-of-the-art algorithms. BoW3D shows better performance in terms of $F_1$ max and extended precision scores on most scenarios. It is noticeable that BoW3D takes an average of 48 ms to recognize and correct the loops on KITTI 00 (includes 4K+ 64-ray LiDAR scans), when executed on a notebook with an Intel Core i7 @2.2 GHz processor. We release the implementation of our method here: https://github.com/YungeCui/BoW3D.
\end{abstract}


\section{Introduction}
\IEEEPARstart{O}{ne} of the basic requirements for long-term simultaneous localization and mapping (SLAM) is the fast and robust loop closing. After a long-term operation, standard frame-to-frame point registration algorithms may fail to align the current observation to the revisited places due to the large drift in the pose estimation. When the revisited places can be robustly recognized, loop closure can provide correct data association to eliminate the drifts resulting in more globally consistent estimation. The same methods used for loop detection can be also used for robot relocalization in presence of the tracking failure case. The basic technique for place recognition is to build a database from the images (for visual SLAM) or point clouds (for LiDAR SLAM) collected online by the robot so that the most similar one can be retrieved when a new sensor observation is acquired. In presence of the same scene detected, a loop closure will be performed. Admittedly, the distance-based association can also be used for loop closing with drift in a certain range. However, when applied to long-term large-scale cases, the pose drift accumulates to a large extent that distance-based association fails to find correct correspondences between the current sensor frame and the historical ones. Therefore, a fast and robust loop closing module is essential for SLAM system.

In the field of visual SLAM, many algorithms \cite{cummins2008fab, angeli2008fast, cadena2012robust, galvez2012bags, mur2014fast, mur2015orb, qin2018vins, campos2021orb} use image retrieval techniques. They typically match images by comparing their 2D features via bag-of-words (BoW) \cite{sivic2003video}. BoW achieves very efficient image retrieval, and have promoted the progress of visual SLAM. Unfortunately, in the field of 3D LiDAR SLAM, the irregularity, sparsity and disorder of LiDAR point cloud raise challenges to 3D feature extraction and representation \cite{cui2022link3d, shi2021keypoint}. Building the BoW for 3D features, applying it to real-time loop closing, and correcting the full 6-DoF loop pose in 3D LiDAR SLAM, are still unsolved.

In this paper, we propose a novel loop closing method that builds the bag of words for 3D features extracted from LiDAR point clouds. We use LinK3D \cite{cui2022link3d} as the 3D feature and build the BoW for 3D LiDAR point clouds named BoW3D. To achieve fast retrieval, the hash table is used as the basic structure of the database. The proposed retrieval and update algorithms can be performed online. After finding the candidate frame, we calculate the full 6-DoF loop pose using the accurate point-to-point LinK3D matching results with RANSAC \cite{fischler1981random} and SVD \cite{arun1987least}. Moreover, we compare our BoW3D with state-of-the-art methods on public dataset, and embed it to the LiDAR odometry A-LOAM \cite{zhang2014loam} to evaluate its performance in practice. The experimental results show that our method achieves significant improvements, and has superior real-time performance. 

\begin{figure*}
	\centering
	\includegraphics[width=1\linewidth]{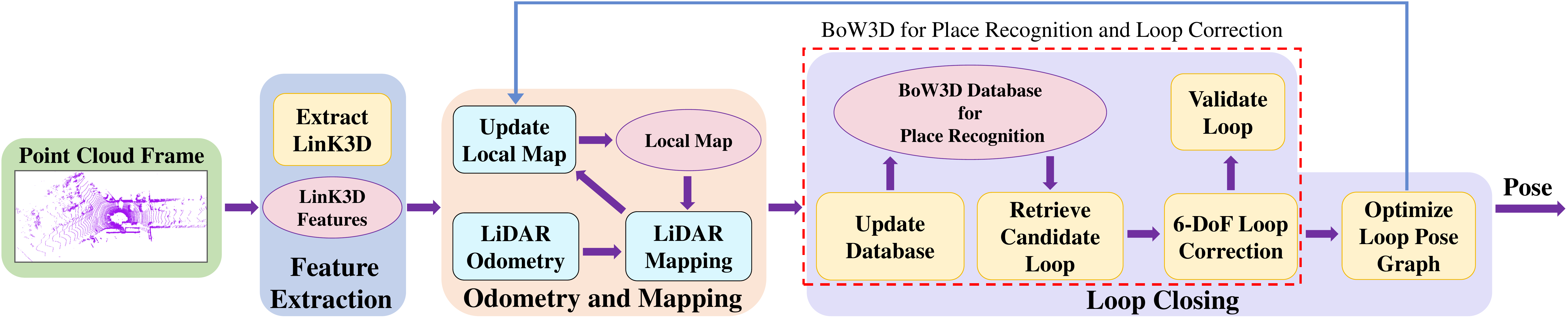}
	\caption{The workflow of our system mainly consists of three modules: (\romannumeral1) Feature Extraction; (\romannumeral2) Odometry and Mapping of A-LOAM; (\romannumeral3) Loop Closing. We embed our BoW3D to the loop closing thread, which closes the loops and obtaines the globally consistent SLAM results.}
\label{fig:fig1}
\end{figure*}

To summarize, our main contributions are as follows:\par
\begin{itemize}
\item We propose a novel BoW-based method using 3D LiDAR features for the loop closing of LiDAR SLAM. It builds the database of LinK3D features, and effectively recognize the revisited places in real time.
\item The proposed BoW3D can also be used to correct the full 6-DoF loop pose in real time, which provides accurate loop closing constraints for subsequent pose graph optimization in online operation.
\item The proposed method has been embedded to the LiDAR odometry system in practice. Experimental results show that our method can significantly eliminate the drifts and improve the accuracy of 3D LiDAR SLAM.
\end{itemize}

\section{Related Work}
Loop closing usually contains two steps. It first finds similar places in the database compared to the current observation, also called place recognition.
Then, it estimates the loop pose and corrects the pose estimation by pose graph optimization. Many different sensor modalities have been used for finding
loops, including camera images and 3D LiDAR points. 

In recent years, a variety of image-based methods \cite{cummins2008fab, galvez2012bags, mur2014fast, arandjelovic2016netvlad, chen2017deep, zaffar2020cohog} have been proposed. A great example is FAB-MAP \cite{cummins2011appearance}, which uses a tree structure to learn the offline words' covisibility probability. However, as relying on SURF \cite{bay2006surf}, it spends a lot of time on feature extraction. DBoW2 \cite{galvez2012bags}, used for the first time bag of binary words based on BRIEF descriptors \cite{calonder2010brief} with the very efficient FAST feature detector \cite{rosten2006machine}, greatly improving the efficiency of loop retrieval. Due to its high efficiency, DBoW2 has been successfully used for loop closing and relocalization in several famous visual SLAM systems \cite{mur2014fast, mur2015orb, forster2016svo, mur2017orb, qin2018vins, campos2021orb}. It is worth noting that, the retrieved feature points based on DBoW2 can also be used for subsequent loop correction and optimization after integrating in long-term visual SLAM. There are three main reasons for the success of applying BoW to visual SLAM: (\romannumeral1) There are many 2D features that have been successfully used in vision tasks, such as SIFT \cite{lowe2004distinctive}, SURF \cite{bay2006surf}, BRIEF \cite{calonder2010brief} and ORB \cite{rublee2011orb}. (\romannumeral2) Camera SLAM usually extracts 2D features, and requires to detect the loop or perform relocalization. As a consequence, bag of words is quite suitable for camera SLAM. (\romannumeral3) Bag of words compresses the image information into a more compact form. Moreover, it builds a tree to speed up the retrieving process of words. Both of these ensure the efficiency of the algorithm in place recognition of camera SLAM.

For methods using 3D point cloud, Steder $et$ $al.$ \cite{steder2011place} propose a place recognition method operating on range images generated from 3D LiDAR data, which uses a combination of bag-of-words and NARF-feature-based \cite{steder2010narf}. M2DP \cite{he2016m2dp} projects a point cloud to multiple 2D planes and generates a density signature for points in each of the planes. The singular value decomposition (SVD) components of the signature are then used to compute a global descriptor. Scan contex \cite{kim2018scan} converts the point cloud scan into a visible space, and uses the similarity score to calculate the distance between two scan contexts. However, this method can only provide relative 1-DoF yaw angle estimation of loop LiDAR pair and fails to correct the full 6-DoF pose of loops.  PointNetVLAD \cite{uy2018pointnetvlad} leverages PointNet \cite{qi2017pointnet} and NetVLAD \cite{arandjelovic2016netvlad} to generate global descriptors. Then it proposes a "lazy triplet and quadruplet" loss function to tackle the retrieval task. ISC \cite{wang2020intensity} proposes a global descriptor based on the geometry and intensity informations, then uses a two-stage hierarchical intensity scan context to detect loops. LiDAR-Iris \cite{wang2020lidar} generates the LiDAR-Iris image representation to detect potential loops. SGPR \cite{kong2020semantic} uses a semantic graph representation for the point cloud scenes by reserving the semantic and topological information of the raw point cloud. OverlapNet \cite{chen2021overlapnet} adopts a siamese network to estimate an overlap of range image generated from LiDAR scans, and provides a relative yaw angle estimate of matching LiDAR pair.  SSC \cite{li2021ssc} uses global semantic scan context to detect loops. OverlapTransformer \cite{ma2022overlaptransformer} uses a lightweight neural network expoiting the range image representation of point cloud to achieve fast retrieval. Overall, most existing methods are either too time-consuming to detect loops in real time for online 3D LiDAR SLAM or can not provide a 6-DoF pose estimation required for LiDAR loop closing.

\section{Background Review}

\subsection{Review of LinK3D Features} \label{review of LinK3D}
\begin{figure}
	\centering
	\includegraphics[width=0.98\linewidth]{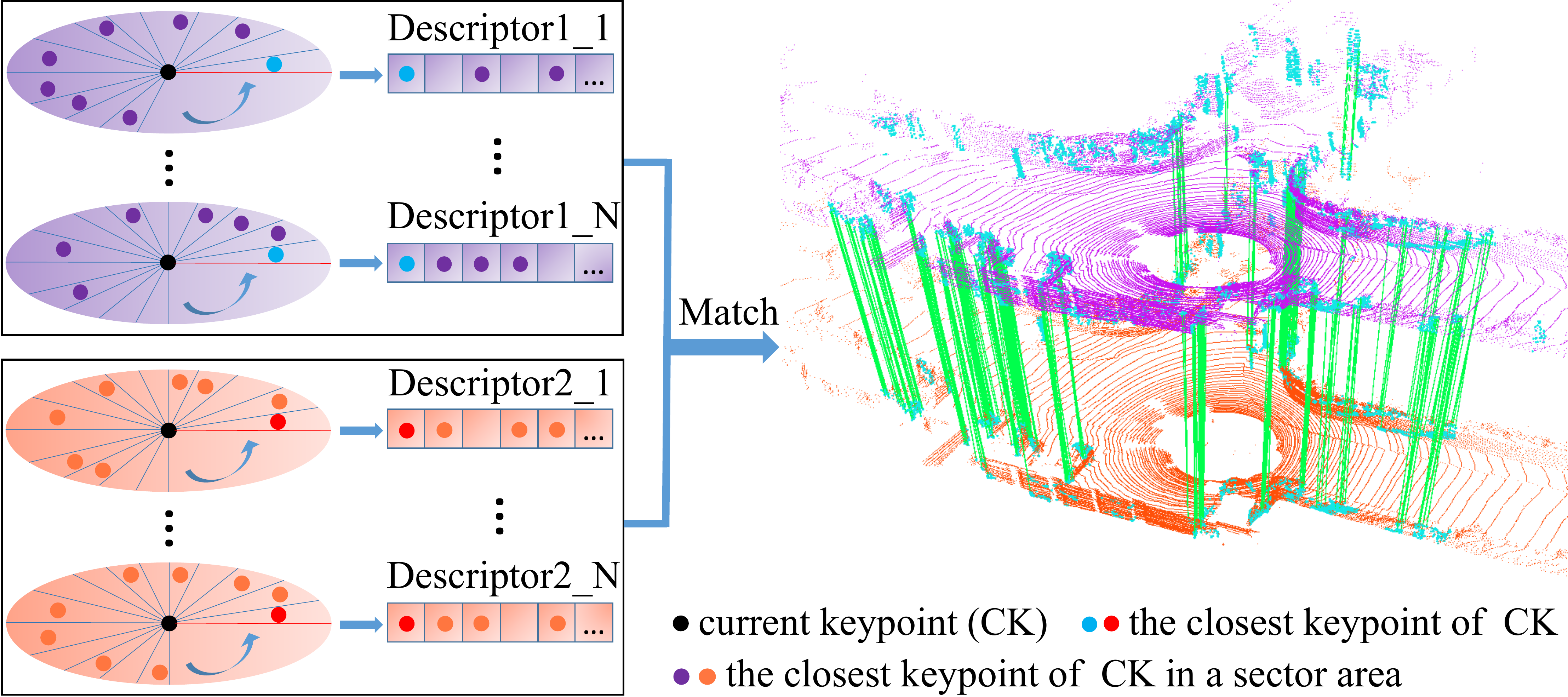}
	\caption{The core idea of LinK3D and the matching result of two LiDAR scans based on LinK3D. The green lines are the valid matches. The descriptor of current keypoint is represented with its neighbor keypoints and described as a multi-dimensional vector. Each dimension of the descriptor corresponds to a sector area. The first dimension corresponds to the sector area where the closest keypoint of current keypoint located, and the others correspond to the areas arranged in a counterclockwise order. If there are keypoints in a sector area, the closest keypoint in the sector area is searched and used to represent the corresponding dimension of the descriptor.}
	\label{fig:fig2}
\end{figure}

In this work, the proposed BoW3D is based on the LinK3D \cite{cui2022link3d} feature. LinK3D consists of three parts: keypoint extraction, descriptor generation and feature matching. As shown in \cref{fig:fig2}, the core idea of LinK3D descriptor is to represent the current keypoint using the neighborhood information, which is inspired by the 2D image features SIFT \cite{lowe2004distinctive} and ORB \cite{rublee2011orb}. The LinK3D descriptor is represented by a 180-dimension vector. Each dimension of the descriptor corresponds to a sector area. The first dimension corresponds to the sector area where the closest keypoint located, and the others correspond to the areas arranged in a counterclockwise order. LinK3D is lightweight and takes an average of 32 ms to extract features from the point cloud collected by a 64-ray laser beam, when executed on a notebook with an Intel Core i7 @2.2 GHz processor. What's more, LinK3D can be used to achieve accurate point-to-point matching, which enables it to be applied to fast 3D registration. For more details about LinK3D, please refer to its original paper \cite{cui2022link3d}.

\subsection{Review of Bag of Words}
In the field of 2D image, bag of words (BoW) \cite{galvez2012bags} is used to recognize the revisited place by retrieving the 2D features (such as SIFT \cite{lowe2004distinctive}, ORB \cite{rublee2011orb}, etc.). In particular, the database of BoW for binary ORB feature has successfully applied to real-time place recognition task. BoW creates a visual vocabulary as a tree structure, in an offline step over a set of descriptors extracted from a training image dataset. When processing a new image, the vocabulary converts the extracted features of the image into a low-dimensional vector, containing the term frequency and inverse document frequency (tf-idf) \cite{sivic2003video} score. The tf-idf score is computed by:
\begin{equation}
	tf \mbox{-} idf = \frac{n_{wi}}{n_i} log\frac{N}{n_w},
\end{equation}
where the $n_{wi}$ represents the number of word $w$ in image $I_i$, $n_i$ represents the total number of words in $I_i$, $N$ is the total number of images seen so far, and $n_w$ represents the number of images containing word $w$. The idea of idf is that the less times a word appears in all images, the higher the final score will be, which indicates that the word has a higher discrimination. The higher of the total tf-idf score, the more frequent of the word in the image. If the score of a word is high enough, the similarity between the word and the words in the database is computed. If there are similar words in the database, then the invert index is used to search the corresponding images.\par

\section{Methodology}
In this section, we introduce the proposed loop closing system based on our BoW3D. To verify the performance of BoW3D in practice, the loop closure system is embedded to the state-of-the-art A-LOAM\footnotemark[6]\footnotetext[6]{https://github.com/HKUST-Aerial-Robotics/A-LOAM} \cite{zhang2014loam}. As shown in Fig. \ref{fig:fig1}, the proposed system mainly consists of three parts. It first extracts the LinK3D features in the raw point clouds. Then it uses the odometry and the mapping algorithms of A-LOAM to estimate the robot poses. The third part uses the proposed BoW3D to detect loop closures and corrects the loop poses. If a loop is detected and optimized, the loop closing will give feedback to the mapping algorithm and update the local map, which provides more accurate estimation for subsequent steps. 

\subsection{BoW3D Algorithm} \label{Bag of words with LinK3D}
\begin{figure}
	\centering
	\includegraphics[width=1\linewidth]{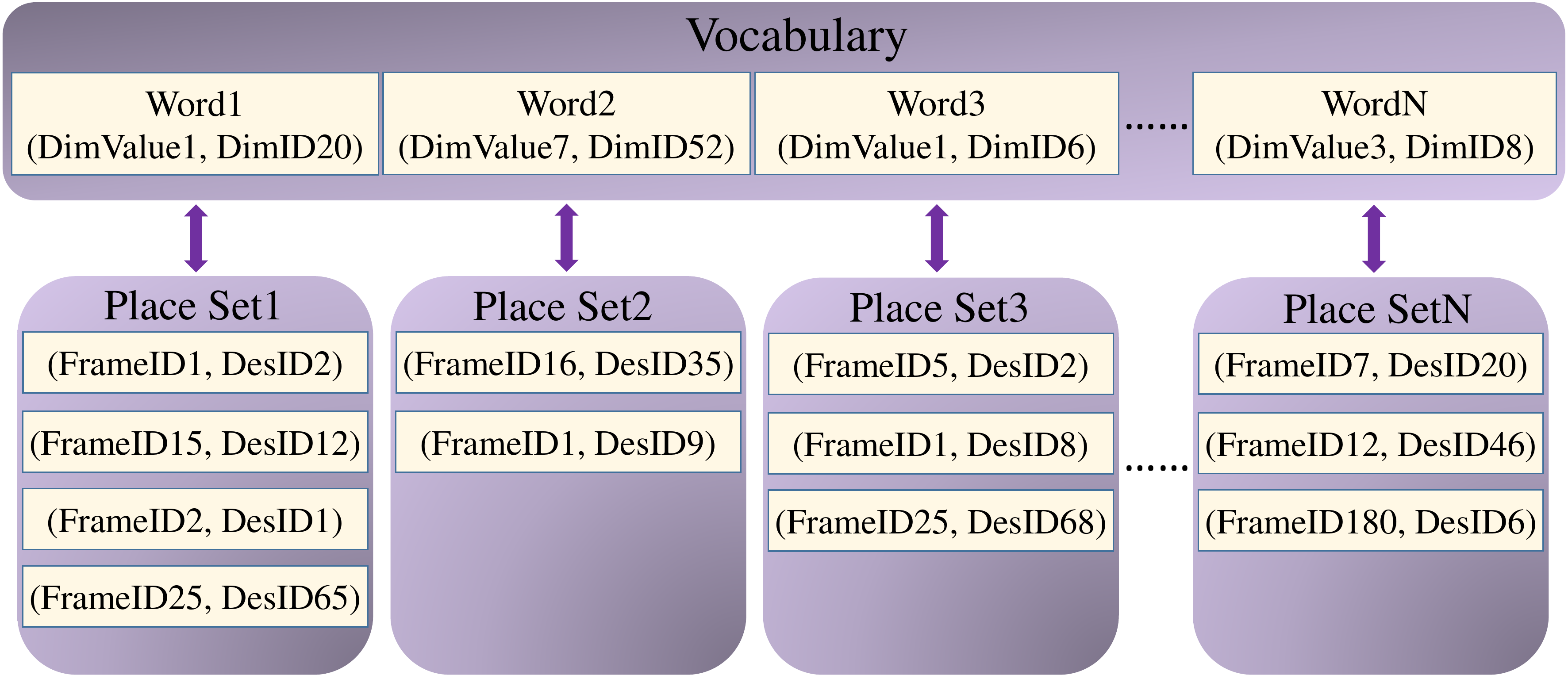}
	\caption{The data structure of BoW3D. The hash table is used for the retrieval. The word of BoW3D consists of the non-zero value (Dim-value) in the descriptor and the corresponding dimension (Dim-ID). Each word corresponds to a place set, in which the word has appeared. The place also consists of two parts, one is frame ID, the other is the descriptor ID in the frame.}
	\label{fig:fig4}
\end{figure}

In this section, we introduce the proposed BoW3D algorithm. As introduced in \cref{review of LinK3D}, each dimension of LinK3D descriptor represents a specific keypoint in the corresponding area, which makes LinK3D descriptor quite discriminative. As a result, our method needs no further conversion for features by building a tree-structured vocabulary in an offline step. Therefore, our BoW3D does not require to load additional vocabulary file, which is more convenient for users. The structure of the database in memory is shown in Fig. \ref{fig:fig4}, the hash table is used to build a one-to-one mapping between the word and the places where the word has appeared. The computational cost of hash table is $O(1)$ theoretically, which makes it be suitable for efficient retrieval. As shown in Fig. \ref{fig:fig4}, the word in the vocabulary of BoW3D is constructed by two parts: One is the non-zero dimension of LinK3D; The other is the corresponding dimension where the word located. The place (point cloud frame) also consists of two parts: One is the frame Id; The other is the descriptor ID in the frame.

\subsubsection{\textbf{Retrieval Algorithm}} For the words of a LinK3D descriptor, we retrieve the words and count the frequency of each place (point cloud frame) appeared. If the highest one is greater than the frequency threshold $Th_f$, this place is considered as a candidate place. In addition, for fast retrieval, the inverse document frequency (idf) is used to avoid retrieving the word appeared in multiple places. More specifically, for the words appears obviously more often than other words, they are less discriminative and reduce the retrieval efficiency. Therefore, we define a $ratio$ factor that is similar to idf to measure the difference between the number of places in current set and the average in all sets. We use it to determine whether we should reserve the place set of current word when count the number of places. The $ratio$ is defined as follows:

\begin{equation} \label{equation2}
ratio = N_{set}/(\frac{N}{n_w}),
\end{equation}
where $N_{set}$ is the number of places in current set. $\frac{N}{n_w}$ represents the quantity average of one place set, $n_w$ is the number of words in vocabulary and $N$ is the total number of places seen so far. If the $ratio$ of a place set is greater than threshold $Th_r$, the place set will not be used for counting. The retrieval algorithm is shown in Algorithm \ref{alg:A}. \par

\subsubsection{\textbf{Loop Correction}} 
Loop correction is used to provide constraint for the pose graph optimization in backend. The observation constraint of the loop is firstly calculated based on the matching result of LinK3D, then RANSAC \cite{fischler1981random} is used to remove the mismatches. We follow the approach in \cite{arun1987least} to calculate the loop pose $\bm{T_{l,c}}$, which is the estimation between the loop frame $l$ and the current frame $c$.\par

Given the point set $\left\{ S \right\}_{c}$ of current frame and the matching point set  $\left \{ S \right\}_l$ of loop frame, we compute the loop by minimizing the following cost function:
\begin{equation}
	r_{l,c}(\bm{R_{l,c}}, \bm{t_{l,c}}) = \frac{1}{2} \sum\limits_{i=1}^{n} \Vert \bm{s}_l^{i} - \left( \bm{R_{l,c}}\bm{s}_c^{i} + \bm{t_{l,c}} \right)  \Vert^2,
\end{equation}
where $\bm{s}_l^{i} \in \left \{ S \right\}_l$ and $\bm{s}_c^{i} \in \left\{ S \right\}_{c}$ are the corresponding matching points. $\bm{R_{l,c}}$ and $\bm{t_{l,c}}$ are the rotation and the translation of $\bm{T_{l,c}}$ transformation respectively. $\bm{T_{l,c}}$ is defined as follows:

\begin{equation}
	\bm{T_{l,c}} = \begin{bmatrix} \bm{R_{l,c}} \hspace{0.8em} \bm{t_{l,c}} \vspace{1.2em} \\  \bm{0}^T \hspace{1.2em} 1 \end{bmatrix} \label{9}.
\end{equation}

We first calculate the centroid $\bm{s_l}$ and $\bm{s_c}$ from $\left \{ S \right\}_l$ and $\left\{ S \right\}_{c}$. Let $\bm{\hat s}_l^{i}$ and $\bm{\hat s}_c^{i}$ be the coordinates of the corresponding point $\bm{s}_l^{i}$ and $\bm{s}_c^{i}$, which removes the centriod $\bm{s_l}$ and $\bm{s_c}$ respectively. Then we calculate the matrix:
\begin{equation}
	\bm{W} = \sum\limits_{i=1}^{n} \bm{\hat s}_l^{i} \bm{\hat s}_c^{iT}.
\end{equation}

Find the SVD decomposition of $\bm{W}$,
\begin{equation}
	\bm{W} = \bm{U} \bm{\Sigma} \bm{V}^T.
\end{equation}

If W is full rank, we will get the solution of $\bm{R_{l,c}}$ and $\bm{t_{l,c}}$ by:
\begin{equation}
	\begin{split}
		\bm{R_{l,c}} = \bm{V U}^T, \\
		\bm{t_{l,c}} = \bm{s_l} - \bm{R_{l,c}} \bm{s_c}.
	\end{split}
\end{equation}

This allows us to correct the loop pose. Moreover, the loop pose is also used for the geometry verification of candidate loop frame, and we set distance threshold $Th_{dis}$ to verify wether the candidate loop is valid.\par

\subsubsection{\textbf{Update Algorithm}} We propose an update algorithm to add new words and places to the database. To improve the efficiency of update and retrieval, we only add a certain number of features to the database. The descriptors are selected based on the distance between their corresponding keypoints and the LiDAR centers. Specifically, only a certain number of closer features are added to the database. Similarly, when retrieve the loops, a certain number of closer features are used to retrieve. The update algorithm is shown in Algorithm \ref{alg:B}.\par

\begin{algorithm}[h!]
\IncMargin{1em}
\caption{Retrieval Algorithm}
\label{alg:A}
\SetKwInOut{Input}{\textbf{Input}}\SetKwInOut{Output}{\textbf{Output}}

\Input{
	\textbf{Des} (LinK3D descriptor of a frame)}
\Output{
	\textbf{Candidate Loop Frame ID}
}
\BlankLine

\textbf{Define:}
\\
$Tab_h$: the hash table used to count how many times of each place occurs\\
$MaxFreqPlace$: the place with the maximum frequency in $Tab_h$\\

\textbf{Main Loop:}
\\ 
\For{a $Word$ in \textbf{Des}}{
	\uIf{the $Word$ is in \textbf{Vocabulary}}{
	$PlaceSet$ = GetPlaceSet($Word$) in \textbf{Database}\;
	$ratio$ = ComputeRatio($PlaceSet$)\;
	\uIf{$ratio > Th_r$}{
		continue next cycle\;
	}
	\Else{
		\For{a $Place$ in $PlaceSet$}{
			
			\uIf{the $Place$ appear in $Tab_h$}{
				The corresponding frequency in $Tab_h$ + 1;
			}
			\Else{Push the $Place$ into  $Tab_h$\;}
		}
	}
}
\Else{continue next cycle;}

}

$MaxFreqPlace$ =	GetPlaceWithMaxFrequency($Tab_h$)\;
\uIf{the maximum frequency \textgreater \ $Th_f$}{
	\textbf{Candidate Loop Frame ID} = Frame ID in $maxFreqPlace$ \;
	return \textbf{Candidate Loop Frame ID}\;
}
\Else{return -1;}

\end{algorithm}

\begin{algorithm}[h!]
\IncMargin{1em}
\caption{Update Algorithm}
\label{alg:B}
\SetKwInOut{Input}{\textbf{Input}}

\Input{
	\textbf{Des} (LinK3D descriptor of a frame)}

\textbf{Main Loop:}
\\
\For{a $Word$ in \textbf{Des}}{
	\uIf{the Word is in \textbf{Vocabulary}}{
		$PlaceSet$ = GetPlaceSet($Word$) in \textbf{Database}\;
		$PlaceSet$.insert($(FrameID, DesID)$);
	}
	\Else{
		$PlaceSet$ = DefineNewSet($(Frame ID, DesID)$)\;
		\textbf{Database}.insertNewWordSet($Word$, $PlaceSet$);
	}
}

\end{algorithm}

\subsection{Loop Optimization}
After correcting the loop pose, the pose graph for the loop frames is built for subsequent loop optimization. The vertexs of pose graph are the global poses to be optimized. The edges (connections between the vertexs) of pose graph are the observation constraints, which consist of the relative poses between the sequential frames, and the corrected loop poses. We define the residual between frame $i$ and $j$ as:

\begin{equation}
r_{i,j}(\bm{T}_{w,i}, \bm{T}_{w,i}) =  \ln (\bm{T}_{i,j}^{-1} \bm{T}_{w,i}^{-1} \bm{T}_{w,j})^{\vee}. \\
\label{}
\end{equation}

The pose graph is optimized by minimizing the following cost function:

\begin{equation}
\begin{aligned}
	\min\limits_T  \left\{ \sum\limits_{(i, j) \in S} \Vert r_{i,j} \Vert^2 + \sum\limits_{(i, j) \in L} \Vert r_{i, j} \Vert^2  \right\}, \\
\end{aligned}
\end{equation}
where $S$ is the set of all sequential edges and $L$ is the set of all loop closure edges. We use the Levenberg-Marquadt method implemented in the graph optimizer g2o \cite{kummerle2011g} to solve the optimization.

After poses have been optimized, we update the local map of mapping thread. Specifically, the points in local map are updated based on the optimized global poses, which are more accurate. This can ensure the global consistency of subsequent estimations.\par

\section{Experiments}
To comprehensively evaluate the performance of our algorithm, we perform experimental evaluation from four aspects: (\romannumeral1) Evaluate place recognition performance; (\romannumeral2) Evaluate the performance of our method when integrated in 3D LiDAR SLAM; (\romannumeral3) Test the hyperparameters and analyze the robustness of our method; (\romannumeral4) Test the runtime for each part of the system. We perform experimental verification on KITTI \cite{geiger2012we} dataset which contains data from different street environments ranging from inner cities to suburb areas. The point clouds in KITTI were collected by a Velodyne HDL-64E S2 at the sensor rate of 10 Hz. There are 11 sequences (i.e., from 00 to 10) with ground truth poses, which are used to determine the number of true loops. Furthermore according to \cite{lu2019deepvcp}, some ground truth poses in KITTI involve large errors, and we do also find that the ground truth pose of sequence 02 and 08 involve large errors in experiment. Therefore, we use the more consistent poses refined by ICP to determine if there is a loop closure in sequence 02 and 08. Similar to SGPR \cite{kong2020semantic}, two point cloud scenes are regarded as a true positive loop pair if the Euclidean distance between them is less than 3 m and the time difference is greater than 30 s. These sequences (00, 02, 05, 06, 07 and 08) with loop closures are selected for the evaluation. We set $Th_{dis} = 3$ in our experiments. Experiments are performed on a notebook with an Intel Core i7 @2.2 GHz processor and 16 GB RAM.

\begin{figure}
	\centering
	\includegraphics[width=0.8985\linewidth]{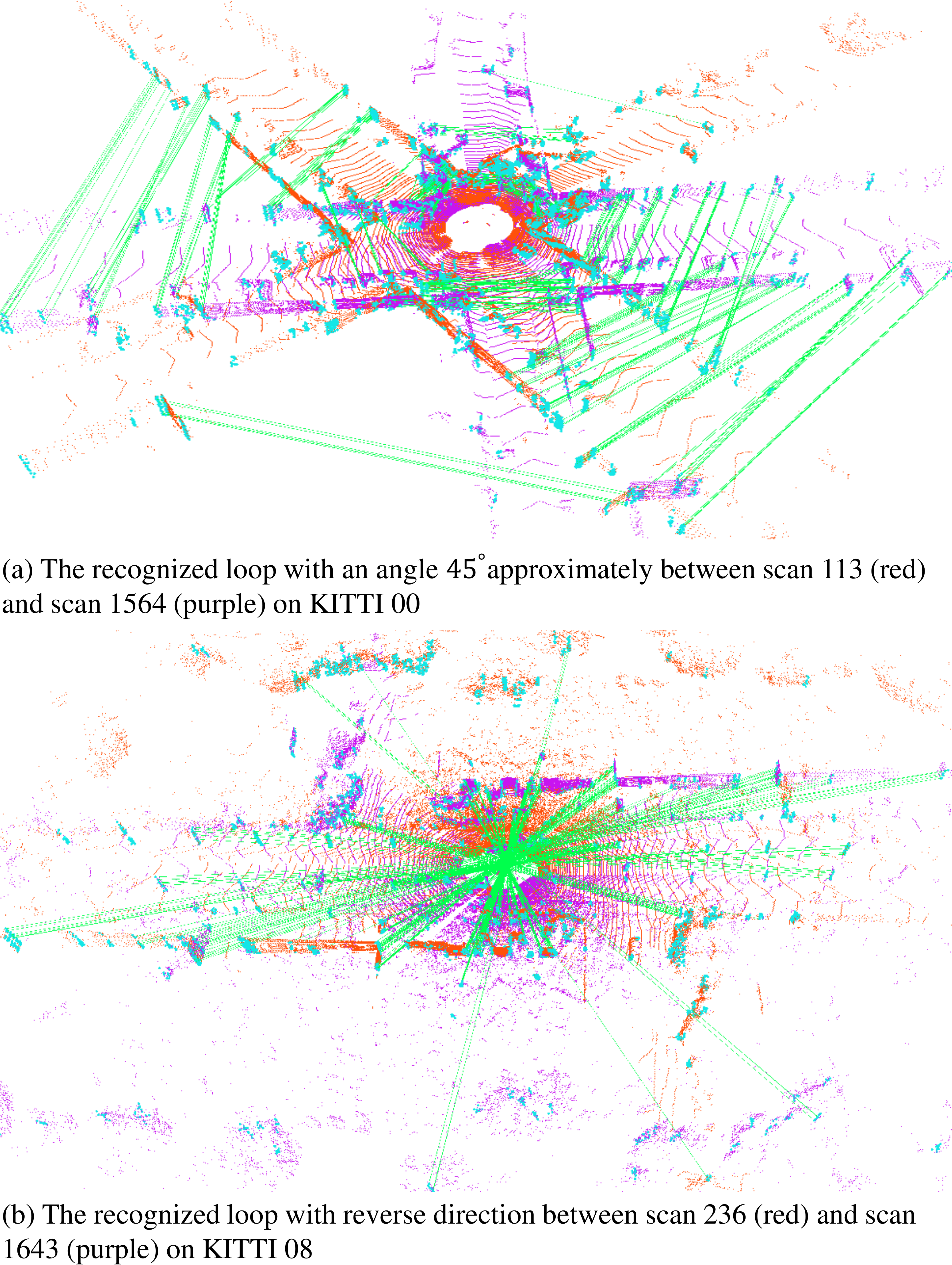}
	\caption{The recognition results based on our place recognition system and the matching results based on LinK3D on loops with different directions. The green lines are the valid matches.}
\label{fig:recognition}
\end{figure}

\subsection{Place Recognition Performance}
In this section, we compare our method with the state-of-the-art LiDAR loop closure detection and place recognition methods, including M2DP \cite{he2016m2dp}, Scan context \cite{kim2018scan}, PointNetVLAD \cite{uy2018pointnetvlad}, Intensity Scan Context (ISC) \cite{wang2020intensity}, LiDAR Iris \cite{wang2020lidar}, SGPR \cite{kong2020semantic}, OverlapNet \cite{chen2021overlapnet} and SSC \cite{li2021ssc}. Following the evaluation metrics introduced in \cite{li2021ssc}, we compute the maximum value of $F_1$ score and Extended Precision \cite{ferrarini2020exploring} (EP) of our method, and adopt the results about $F_1$ score and EP presented in \cite{li2021ssc} for comparison methods. In addition, we also show whether these methods can be used to correct the full 6-DoF loop pose. The $F_1$ score is defined as follows: 

\begin{equation}
	\begin{aligned}
		F_1 = 2 \times \frac{P \times R}{P + R}, \\
	\end{aligned}
\end{equation}
where $P$ and $R$ are the precision and recall, respectively. $F_1$ represents the harmonic mean of precision and recall, which treats precision and recall as equally important and is used to evaluate the overall performance of classification. We use the maximum $F_1$ score of each method for the comparison.

The Extended Precision is defined as follows:
\begin{equation}
	\begin{aligned}
		EP = \frac{1}{2}(P_{R0} + R_{P100}), \\
	\end{aligned}
\end{equation}
where $P_{R0}$ represents the precision at minimum recall, and $R_{P100}$ represents the max recall at $100\%$ precision. EP is specifically designed to validate the place recognition performance. The comparison results are shown in \cref{tab:Recall}. \par

\begin{figure*}
	\centering
	\includegraphics[width=0.9\linewidth]{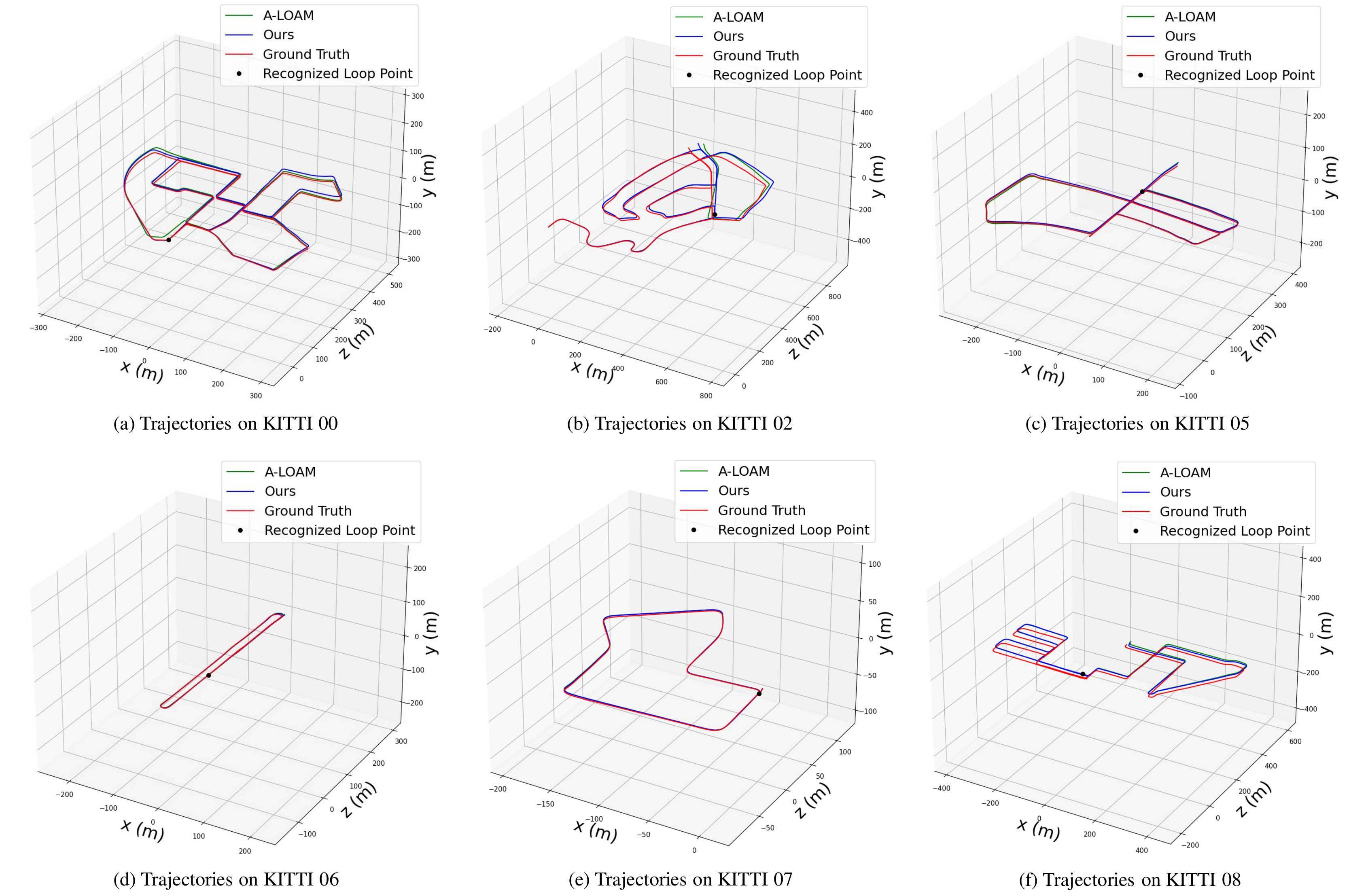}
	\caption{Comparisons between the trajectories of A-LOAM and the trajectories with our place recognition system on KITTI. We can see that our place recognition system can effectively recognize the revisited place and correct the loop pose.}
\label{fig:fig9}
\end{figure*}

\begin{table*}[!ht]
\
\begin{center}
	\caption{The comparison results about $F_1$ max scores, Extended Precision ($F_1$ max scores / Extended Precision) and whether these methods can be used to correct the full 6-DoF loop pose. The best scores are marked in bold and the second best scores are underlined.}
	\footnotesize
	\begin{tabular}{  c | c | c | c | c | c | c | c }
		\toprule
		Method & 00 & 02 & 05 & 06 & 07 & 08 & loop correction \\
		\midrule		
		M2DP \cite{he2016m2dp} & 0.708/0.616 & 0.717/0.603 & 0.602/0.611 & 0.787/0.681 & 0.560/0.586 & 0.073/0.500 & -\\
		PointNetVLAD \cite{uy2018pointnetvlad} & 0.779/0.641 & 0.727/0.691 & 0.541/0.536 & 0.852/0.767 & 0.631/0.591 & 0.037/0.500 & -\\
		ISC \cite{wang2020intensity} & 0.657/0.627 & 0.705/0.613 & 0.771/0.727 & 0.842/0.816 & 0.636/0.638 & 0.408/0.543 & -\\
		LiDAR Iris \cite{wang2020lidar} & 0.668/0.626 & 0.762/0.666 & 0.768/0.747 & 0.913/0.791 & 0.629/0.651 & 0.478/0.562 & -\\
		SGPR \cite{kong2020semantic} & 0.820/0.500 & 0.751/0.500 & 0.751/0.531 & 0.655/0.500 & 0.868/0.721 & 0.750/0.520 & -\\
		Scan contex \cite{kim2018scan} & 0.750/0.609 & 0.782/0.632 & 0.895/0.797 & 0.968/0.924 & 0.662/0.554 & 0.607/0.569 & 1-DoF \\
		OverlapNet \cite{chen2021overlapnet} & 0.869/0.555 & \underline{0.827}/0.639 & 0.924/0.796 & 0.930/0.744 & 0.818/0.586 & 0.374/0.500 & 1-DoF\\
		SSC-RN \cite{li2021ssc} & \underline{0.939}/\underline{0.826} & \textbf{0.890}/\textbf{0.745} & \underline{0.941}/\underline{0.900} & \textbf{0.986}/\underline{0.973} & \underline{0.870}/\underline{0.773} & \underline{0.881}/\underline{0.732} & 3-DoF\\
		\midrule 					
		BoW3D (ours)  & \textbf{0.977}/\textbf{0.981}  & 0.578/\underline{0.704} & \textbf{0.965}/\textbf{0.969} & \underline{0.985}/\textbf{0.985} & \textbf{0.906}/\textbf{0.929} & \textbf{0.900}/\textbf{0.866} & \textbf{6-DoF}\\			
		\bottomrule
		
	\end{tabular}
	\label{tab:Recall}
\end{center}
\end{table*}

As shown in \cref{tab:Recall}, our BoW3D surpasses other methods in $F_1$ max score and $EP$ on most sequences. Especially on sequence 08 with only reverse loops, our method can still retrieve most loops and also achieve high precision. This indicates that our method is robust to the change of view angle, and \cref{fig:recognition} shows the situations. In addition, the comparison methods cannot be used to correct the full 6-DoF loop pose, which limits subsequent loop optimization in 3D LiDAR SLAM. Our method performs well while also can be used to correct the full 6-DoF loop pose. The reasons why our method outperforms the baseline methods are as follows: (\romannumeral1) Our BoW3D is based on the LinK3D descriptors, and LinK3D is pose invariant. This allows our method quickly recognize the loops, even if the loops are reverse. (\romannumeral2) As shown in Fig. \ref{fig:recognition}, our system forms the constraints based on the accurate point-to-point matching results, and this ensures that the effective loop correction can be performed.

\subsection{Performance on LiDAR-based SLAM}
In this experiment, we evaluate the performance of the loop closing system used in 3D LiDAR-based SLAM. To verify the accuracy of loop correction, we refer to the validation metrics defined in \cite{lu2019deepvcp}: (\romannumeral1) Euclidean distance for translation; (\romannumeral2) $\Theta$ for rotation. $\Theta$ is defined as follows:

\begin{equation}
	\label{equation5}
	\begin{aligned}
		\Theta = 2 \arcsin(\frac{\Vert R - \overline{R} \Vert_F}{\sqrt{8}}),
	\end{aligned}
\end{equation}
where $\Vert R - \overline{R} \Vert_F$ is the Frobenius norm of the chordal distance \cite{hartley2013rotation} between estimation and ground truth rotation matrix. 

To verify the accuracy of the whole trajectories, the root mean square error ($RMSE$) is used for the evaluation. $RMSE$ is defined as follows: 

\begin{equation}
	\label{equation6}
	\begin{aligned}
		RMSE = (\frac{1}{m} \sum_{i=1}^{m} {\Vert trans(Q_i^{-1}P_i)\Vert}^2)^{\frac{1}{2}},
	\end{aligned}
\end{equation}
where $Q_i$ is the ground truth pose and $P_i$ is the estimated pose. The loop correction accuracy and the trajectory comparison results between our SLAM system and A-LOAM are shown in \cref{tab:Accuracy} and \cref{fig:fig9}, respectively.

From the comparison results in \cref{fig:fig9}, we can see that the loop closing system can effectively correct the cumulative errors, and the pose estimations with loop closing are better than the original ones. We show the quantitative decrease of $RMSE$ after using our loop closing compared to the original SLAM pose estimation results in \cref{tab:Accuracy}. The results demonstrate that the proposed loop closing can effectively reduce the drifts of 3D LiDAR SLAM system.

\begin{table}[ht]
	\
	\begin{center}
		\caption{The results of angular, translation errors and the variation of RMSE based on the ground truth when BoW3D used to loop correction and optimization.}
		\footnotesize
		\begin{tabular}{ c | c c c c c c }
			\toprule
			Error & 00 & 02 & 05 & 06 & 07 & 08\\
			\midrule
			Angular Err ($^{\circ}$) & 0.685 & 1.130 & 0.598 & 0.289 & 0.532 & 1.480 \\
			Transl. Err ($m$) & 0.764 & 0.162 & 0.238 & 0.060 & 0.138 & 0.037\\
			\midrule 
			$RMSE$ $(m)$ $(\downarrow)$ & 1.069 & 3.173 & 0.958 & 0.542 & 0.153 & 3.945\\
			\bottomrule		
		\end{tabular}
		
		\label{tab:Accuracy}
	\end{center}
	\vspace{-0.3cm}
\end{table}

\begin{figure}
	\centering    
	\includegraphics[width=0.95\linewidth]{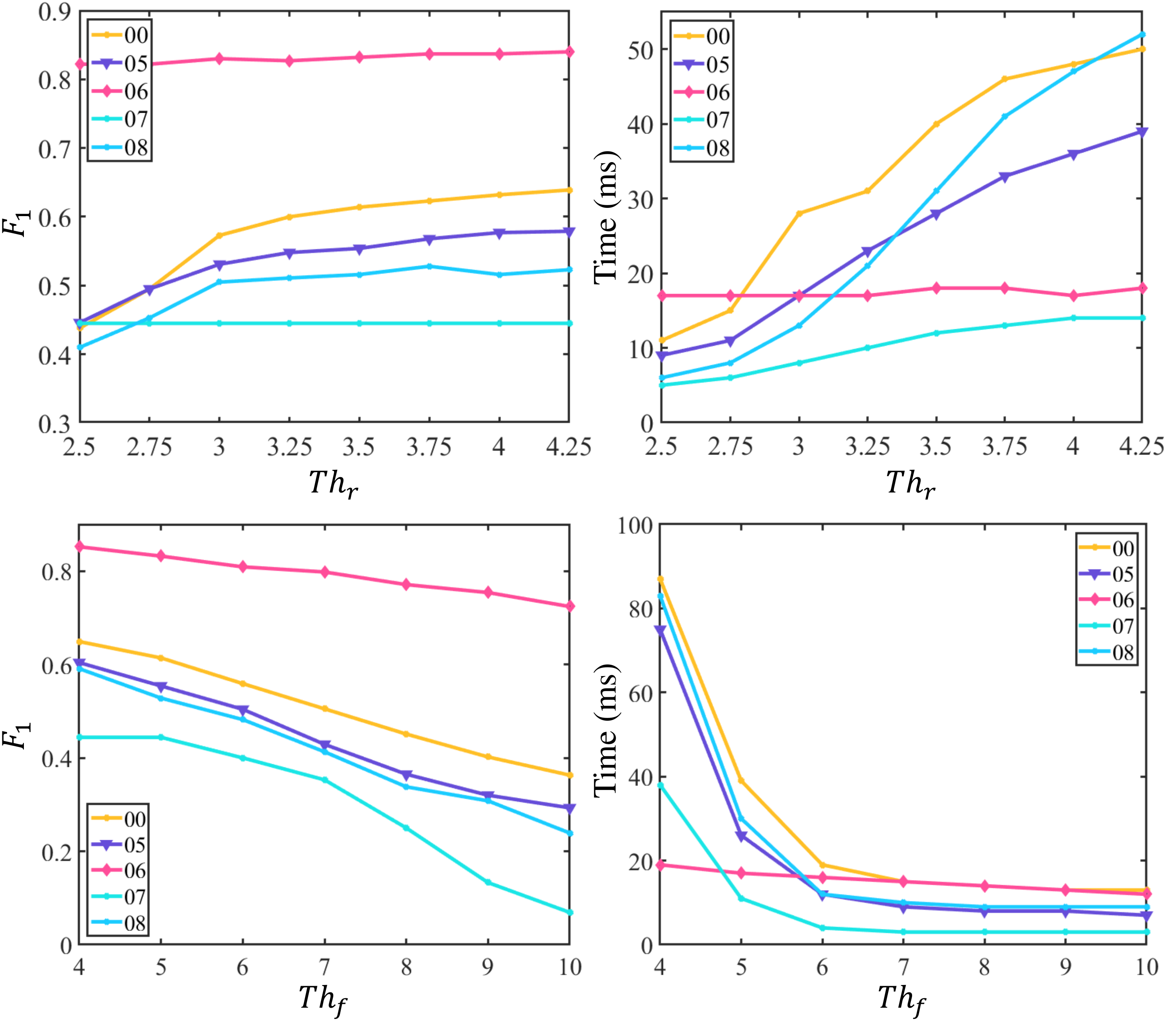}
	\caption{The results of $F_1$ score and runtime with different $Th_r$ and $Th_f$ settings on KITTI 00.}
	\label{fig:fig5}
	\vspace{-0.3cm}
\end{figure}

\subsection{Hyperparameters Setup and Robustness Analyzation} \label{Tests on the parameter}
In this section, we provide more studies on parameter settings about $Th_r$ and $Th_f$ in Algorithm \ref{alg:A} through experiment, which is important to the performance of BoW3D. For a wide test, we set eight parameters about $Th_r$, and it starts from 2.5 to 4.25 with an interval of 0.25. $Th_f$ starts from 4 to 10 with an interval of 1. The $F_1$ score and the average detection time of BoW3D are used to measure the performance of different parameter settings. The results are shown in Fig. \ref{fig:fig5}.

\textbf{Robustness Analyzation}. We can see from Fig. \ref{fig:fig5} that as $Th_r$ increases, the required runtime also increases and the same as the $F_1$ scores. When we set $Th_r$ larger than 3, the $F_1$ scores not change much on most sequences and the runtime still increases. When we set $Th_f$ less than 5, although the $F_1$ is higher, the algorithm also requires much more time to detect the true places. If we set $Th_f$ larger than 8, this will reduce the robustness of our algorithm because of the low $F_1$ score. To make our algorithm robustly work, and trade off the runtime and accuracy, we set the $Th_r$ as 4 and the $Th_f$ as 5.

\begin{figure}	
	\centering
	\includegraphics[width=0.8\linewidth]{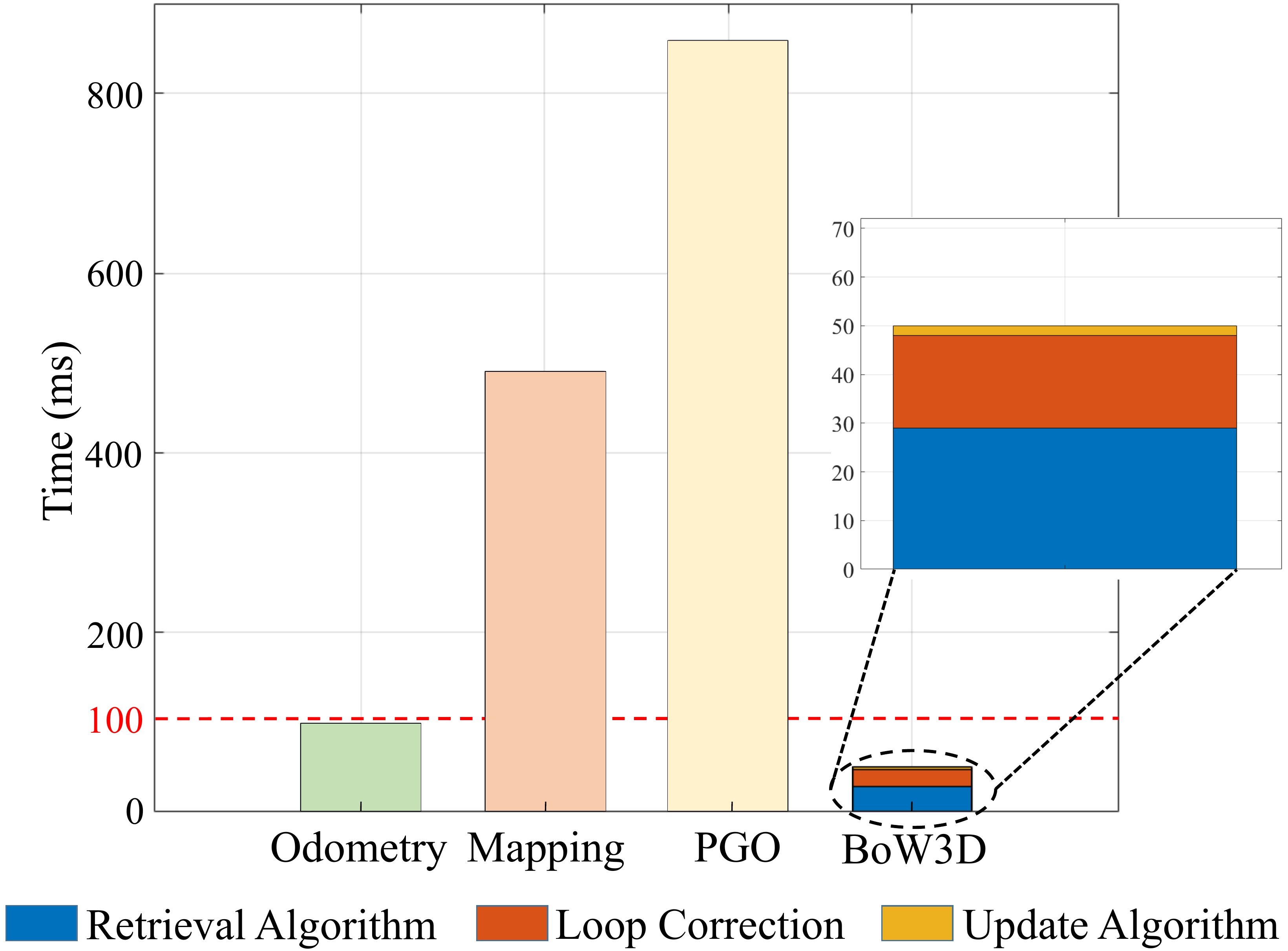}	
	\caption{The average runtime for each part of the system to process one point cloud frame on KITTI 00.}   
	\vspace{-0.3cm} 
	\label{fig:fig6}
\end{figure}

\subsection{System Runtime}
In this experiment, we evaluate the average runtime of each module in the SLAM system after integrating our loop closing for processing one LiDAR scan. KITTI 00 is used for the evaluation, which includes 4K+ LiDAR scans. We set the $Th_r = 4$, $Th_f = 5$. Moreover, we set the number of closer features as 5 when add them to the database, and set it as 3 when retrieve from the database. The runtime of each module is shown in \cref{fig:fig6}. Note that each module of the system operates separately in different threads. Although the runtime of mapping thread and pose graph optimization (PGO) are more than 100 ms, they can be performed online due to their low frequency. In particular, BoW3D takes overall less than 100 ms to process one frame, which ensures the real-time performance of the system when BoW3D applied to 3D LiDAR SLAM.

\section{Conclusion}
In this paper, we propose a novel 3D-feature-based bag of words algorithm for place recognition. The proposed BoW3D exploits the LinK3D features to build the bag of words. It consists of three parts, i.e. place retrieval, loop correction and database update. The hash table is used as the overall structure of the database in BoW3D, which enables to retrieve efficiently. Compared with the state-of-the-art methods, BoW3D not only achieves competitive results, but also corrects the full 6-DoF relative loop pose in real time. Compared with deep-learning-based methods, our method does not require pre-training and GPU resources. In addition, we also embed the proposed BoW3D to LiDAR odometry system to verify its performance in practice. Compared with original odometry algorithm, the 3D LiDAR SLAM system with BoW3D has lower drifts when there are loops. It would be interesting to overcome the shortcomings of current 3D LiDAR odometry systems without loop closing thread, and we will also extend our method to relocalization, local optimization and mapping in 3D LiDAR SLAM, to improve the efficiency, accuracy and robustness of current 3D LiDAR SLAM system.

\bibliographystyle{IEEEtran}
\bibliography{IEEEabrv,MyRef}

\end{document}